\documentclass[10pt,twocolumn,letterpaper]{article}

\usepackage{cvpr}              
\usepackage{balance}
\usepackage[table,svgnames]{xcolor}
\usepackage{booktabs}

\usepackage{pifont}
\usepackage{algorithm}
\usepackage{algorithmic}

\definecolor{cvprblue}{rgb}{0.21,0.49,0.74}
\usepackage[pagebackref,breaklinks,colorlinks,allcolors=cvprblue]{hyperref}

\def\confName{CVPR}
\def\confYear{2026}

\newcommand{\methodAbbr}{S\textsuperscript{2}DiT}

\title{
\methodAbbr: \underline{S}andwich Diffusion Transformer for Mobile \underline{S}treaming Video Generation
}

\author{
\begin{tabular}{c}
Lin Zhao$^{1,2,*}$ \quad
Yushu Wu$^{1,2,*}$ \quad
Aleksei Lebedev$^{1}$ \quad
Dishani Lahiri$^{1}$ \quad
Meng Dong$^{1}$ \\
Arpit Sahni$^{1}$ \quad
Michael Vasilkovsky$^{1}$ \quad
Hao Chen$^{1}$ \quad
Ju Hu$^{1}$ \quad
Aliaksandr Siarohin$^{1}$ \\
Sergey Tulyakov$^{1}$ \quad
Yanzhi Wang$^{2}$ \quad
Anil Kag$^{1}$ \quad
Yanyu Li$^{1,\dagger}$
\end{tabular}
\\[6pt]
$^1$Snap Inc. \quad $^2$Northeastern University \\
$^*$ Equal contribution \qquad $^\dagger$ Corresponding author \\
\href{https://snap-research.github.io/S2DiT/}{Project Page}
}

\begin{document}
\twocolumn[{
\renewcommand\twocolumn[1][]{#1}
\maketitle

\begin{center}
    \centering
    \captionsetup{type=figure}
    \includegraphics[width=1.\linewidth]{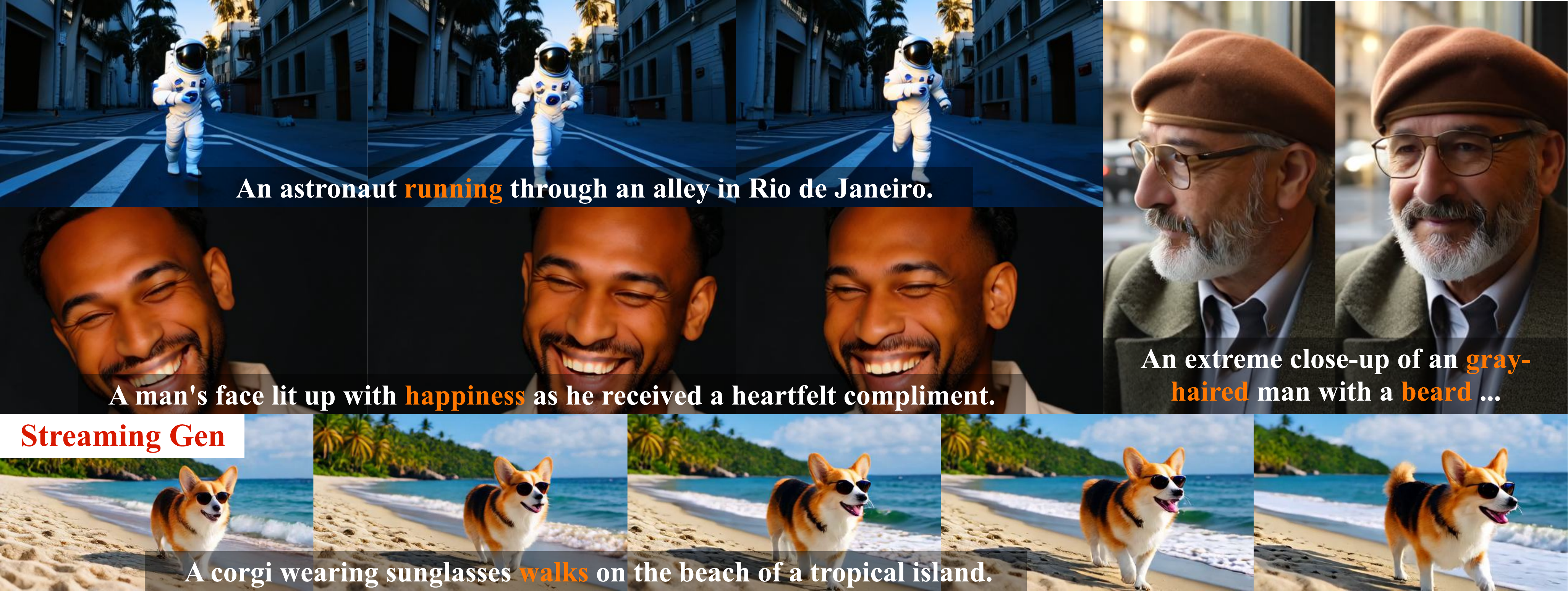}
    \vspace{-2em}
    \caption{\textbf{Example results generated by our \methodAbbr.} Our on-device model can support efficient streaming video generation.}
    \label{fig: teasor}
\end{center}

}]

\begin{abstract}

Diffusion Transformers (DiTs) have recently improved video generation quality.
However, their heavy computational cost makes real-time or on-device generation infeasible. 
In this work, we introduce \methodAbbr—a Streaming Sandwich Diffusion Transformer designed for efficient, high-fidelity, and streaming video generation on mobile hardware. 
\methodAbbr~generates more tokens but maintains efficiency with novel efficient attentions: a mixture of LinConv Hybrid Attention (LCHA) and Stride Self-Attention (SSA).
Based on this, we uncover the sandwich design via a budget-aware dynamic programming search, achieving superior quality and efficiency.
We further propose a 2-in-1 distillation framework that transfers the capacity of large teacher models (e.g., Wan 2.2-14B) to the compact few-step sandwich model. 
Together, \methodAbbr~achieves quality on par with state-of-the-art server video models, while streaming at over 10 FPS on an iPhone.


\end{abstract}    
\section{Introduction}
\label{sec:intro}

Diffusion Transformers (DiTs) \cite{wan2025wan,sun2024hunyuan,hong2022cogvideo,opensora} have rapidly advanced the frontier of video generation, achieving state-of-the-art quality in both foundational text/image to video generation and downstream tasks such as video editing, upsampling and frame interpolation \cite{wu2023tune, zhou2024upscale,wang2024animatelcm,AnimateDiff}.

Despite the rapid progress, several major bottlenecks still remain, particularly in inference speed and resource efficiency.
A key source of the limitations lies in the quadratic computational complexity and memory footprint of the attention mechanism over a large number of tokens, which fundamentally hinder real-time and on-device generation.
Recent works have explored the efficiency of video DiT, \eg, LTX \cite{hacohen2024ltx}, SnapGenV \cite{wu2024snapgenvgeneratingfivesecondvideo}, SnapGenV-DiT \cite{wu2025tamingdiffusiontransformerefficient}, SANA Video \cite{chen2025sanavideoefficientvideogeneration}. 
They mostly rely on high-compression video VAEs to obtain a compact latent space \cite{LTX-video,wu2025tamingdiffusiontransformerefficient}, but struggle to match the visual fidelity and temporal coherence of top‐tier large-scale models due to the significantly reduced token count.
Meanwhile, streaming video generation models~\cite{ yin2025causvid,lin2025autoregressiveadversarialposttrainingrealtime,huang2025selfforcing} have drawn increasing attention because of their on-the-fly and interactive generation capabilities.
Such models impose higher demands on real-time, on-device generation compared to the base bidirectional models \cite{wu2024snapgenvgeneratingfivesecondvideo,wu2025tamingdiffusiontransformerefficient,karnewar2025neodragon}, yet mobile deployment remains largely underexplored.
The open problem is that:
\textit{How to achieve high fidelity, mobile-efficient, and streaming-capable video generation simultaneously?}

In this work, we propose S²DiT, a \textbf{S}andwich diffusion transformer for mobile \textbf{S}treaming video generation as presented in \cref{fig: teasor}. 
We address the challenge from two aspects.

\noindent\textbf{An Efficient Sandwich Diffusion Transformer.}
To overcome the quadratic cost of conventional self-attention in DiTs, we design a multi-stage “sandwich” DiT architecture that interleaves two efficient attention modules. The LinConv Hybrid Attention (LCHA) combines a learnable positive-kernel linear path with a depthwise 3D convolution path, achieving linear complexity while preserving spatiotemporal fidelity. 
The Stride Self-Attention (SSA) compresses intermediate feature maps to improve throughput. 
Based on this, we propose a dynamic programming–based search algorithm that determines the placement of LCHA and SSA modules by optimizing their allocation under latency and memory constraints.
The resulting Sandwich DiT achieves superior quality and speed compared to multiple ablated alternatives, offering a strong backbone for mobile video generation.

\noindent\textbf{2-in-1 Distillation Pipeline. }
Building upon the architecture, we introduce a \textbf{2}-stage distillation pipeline guided by \textbf{1} single teacher model, enabling high-fidelity generation with limited computational budgets. After achieving training convergence, we first perform cached offline knowledge distillation from a large-scale teacher model (i.e., Wan 2.2-14B).
Instead of relying on expensive on-the-fly teacher inference, we precompute and cache diffusion tuples and text embeddings from the teacher models, enabling cost-efficient supervision of the student model. 
This protocol preserves semantic consistency while substantially reducing training FLOPs and peak memory, effectively transferring the visual quality of billion-scale models to a compact mobile backbone.

For the second stage, we extend our framework to streaming video generation through the self-forcing strategy \cite{huang2025selfforcing}.
We employ step distillation to shorten the denoising trajectory, leveraging distribution matching distillation with the same large-scale teacher model used in the first stage.
In addition, we explore adversarial fine-tuning to enforce temporal coherence across streaming segments under a few sampling steps. 
The resulting model can synthesize videos in an online, causal manner, maintaining frame-to-frame consistency while achieving high-speed generation on mobile devices.
To our knowledge, this is the first diffusion transformer enabling on-device streaming video generation with both high fidelity and low latency as shown in \cref{table: compare_intro}. Our contributions can be summarized as follows:
\begin{itemize}[itemindent=0em, ]
    \item We propose \methodAbbr, a streaming sandwich-like diffusion transformer that interleaves two efficient attention modules—hybrid linear-local attention (LCHA) and Stride Self-Attention (SSA), to balance global and local modeling under mobile constraints.
 In addition, we propose a dynamic programming–based search algorithm that automatically allocates LCHA and SSA to optimize the latency–fidelity trade-off.
    \item We introduce a 2-in-1 distillation framework for \methodAbbr~that transfers the high-fidelity generation capability of large-scale teacher models (\eg, Wan 2.2-14B) to the compact, few-step sandwich transformer, yielding an efficient auto-regressive video diffusion model.
    \item We are the first to demonstrate high-fidelity, high-dynamic, and high-speed streaming video generation on a mobile device.
\end{itemize}

\begingroup
\setlength{\textfloatsep}{0pt}
\setlength{\intextsep}{0pt}
\begin{table}[t]
\small
\centering
\resizebox{0.8\linewidth}{!}{
\begin{tabular}{lccc}
\toprule
Model           & VBench & Mobile & Mobile Streaming \\
\hline
Wan2.1-1.3B \cite{wan2025wan}    & 83.31  & \ding{55}      & \ding{55}                \\
LTX-2B    \cite{LTX-video}      & 80.00  & \ding{55}      & \ding{55}                \\
SnapGenV    \cite{wu2024snapgenvgeneratingfivesecondvideo}    & 81.14  & 5s / 4s  & \ding{55}                \\
Mobile-DiT~\cite{wu2025tamingdiffusiontransformerefficient}        & 81.45  & 5s / 4s  & \ding{55}                \\
NeoDragon~\cite{karnewar2025neodragon}     & 81.61  & 2s / 6.7s  & \ding{55}                 \\
\hline
\methodAbbr-AR           & 83.26  & \checkmark  & $\sim11$ FPS           \\
\bottomrule
\end{tabular}
}
\caption{\textbf{We present \methodAbbr, the first-ever mobile streaming video generation model with comparable quality to best server ones.} For methods that support mobile deployment, we report video length / latency. }
\label{table: compare_intro}
\end{table}
\endgroup

\section{Related Works}
\label{sec:related works}

\textbf{Video Diffusion Models.} 
\noindent In recent years, video generation models have advanced rapidly~\citep{wan2025wanopenadvancedlargescale,sora,opensoraplan,CogVideoX,HunyuanVideo,genmo2024mochi,kling,StepVideoT2V}.
Most progress has centered on large diffusion models that iteratively denoise Gaussian noise into real videos conditioned on text or image inputs.
These methods encompass both pixel-space approaches~\citep{SnapVideo,ImagenVideo} and latent-space variants~\citep{wan2025wanopenadvancedlargescale,CogVideoX}.
Although systems such as~\citep{sora,opensora,SnapVideo,MovieGen,LTX-video,CogVideoX} can produce highly realistic videos, their substantial compute and memory footprint limits their practicality for on-device deployment.

\noindent\textbf{Mobile Deployment.}
Relatively few works have explored on-device video generation~\citep{wu2024snapgenvgeneratingfivesecondvideo,kim2025ondevicesoraenablingtrainingfree}. Although Wan2.1~\citep{wan2025wanopenadvancedlargescale} offers $1.3$B T2V model, it has a high number of tokens for on-device inference due to low VAE compression. LTX-Video~\citep{hacohen2024ltxvideorealtimevideolatent} employs a highly compressed VAE with a $1.9$B DiT, although runs in real-time on GPUs, it remains impractical for mobile devices. Mobile Video Diffusion~\citep{yahia2024mobilevideodiffusion} simplifies Stable Video Diffusion~\citep{SVD} by pruning channels and blocks. SnapGen-V~\citep{wu2024snapgenvgeneratingfivesecondvideo} compromises visual quality due to a low-capacity lightweight UNet architecture. On-device Sora~\citep{kim2025ondevicesoraenablingtrainingfree} enables low-resolution video generation on iPhones by merging temporal tokens and dynamically loading blocks to alleviate memory constraints.

\noindent\textbf{Streaming Video Generation} 
Several recent works have specifically tackled the challenge of streaming or causal video generation rather than the conventional full-sequence (bidirectional) diffusion approach.
For example, CausVid \cite{yin2025causvid} introduces an auto-regressive diffusion transformer by converting a pretrained bidirectional backbone into one that generates frames on the fly, achieving real-time streaming~(9.4 FPS) through key-value caching and a distribution-matching distillation from 50 steps to 4 steps. 
Self-forcing \cite{huang2025selfforcing} mitigates exposure bias in auto-regressive video diffusion by simulating inference at training time—performing self-rollouts, conditioning on model-generated context, and applying a holistic video-level loss—thus enabling sub-second latency streaming on a single GPU. 
AAPT \cite{lin2025autoregressiveadversarialposttrainingrealtime} proposes an adversarial post-training method that transforms a pretrained latent video diffusion model into a real-time interactive streamer, enabling one latent frame per network forward evaluation and supporting long-duration (\eg, 1 minute+) streaming at 24 fps with low latency.

\begin{figure*}[t]
    \centering
    \includegraphics[width=1.0\linewidth]{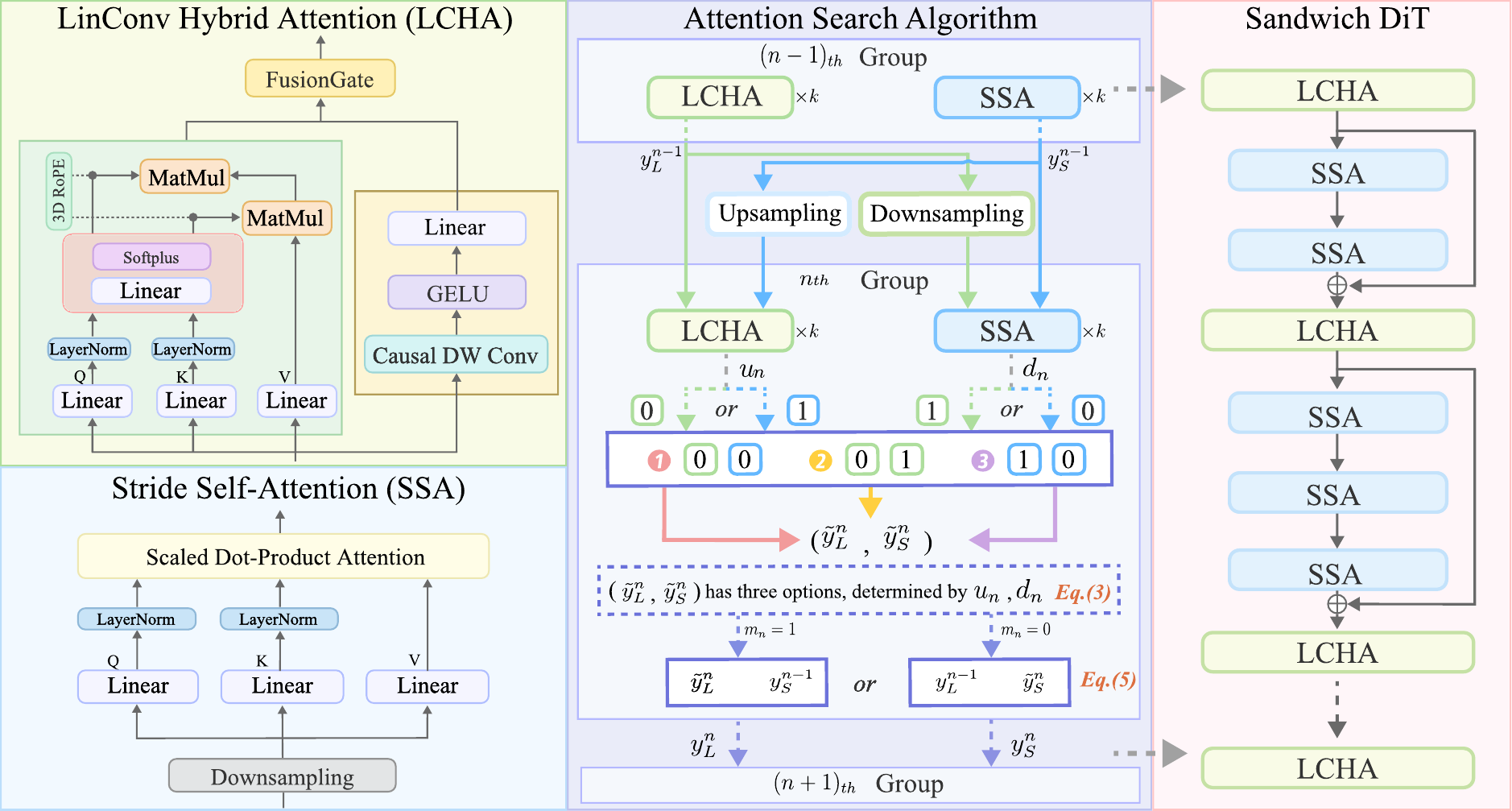}
    \caption{
    \textbf{Illustration of the framework for obtaining~\methodAbbr.}
    LCHA integrates a linear attention path with a local convolution path at high resolution, while SSA compresses the spatial representation for efficient global context modeling.
    The final \methodAbbr~is derived by combining these two efficient attention designs with the attention search algorithm.
    }
    \label{fig:framework}
\end{figure*}

\section{Method}


\subsection{Design Overview}

Unlike prior mobile video diffusion systems that rely on extremely compressed latent spaces, S²DiT operates in a moderately compressed latent representation that preserves more spatial and temporal detail. 
This choice improves fidelity but significantly increases the number of latent tokens, making conventional DiT architectures too slow for mobile deployment. 

Our approach addresses this challenge through a structured architectural pattern that alternates high-resolution and low-resolution processing stages. 
At a high level, Sandwich DiT interleaves two complementary attention modules: LinConv Hybrid Attention (LCHA) for high-resolution modeling that preserves detail at linear cost, and Stride Self-Attention (SSA) for low-resolution modeling that aggregates global context at reduced token count. 
The architectural layout and the precise allocation of these modules are determined through a budget-aware search.

In addition, distillation from a powerful teacher model plays a central role, supplying rich semantic and structural guidance that leads to better visual quality and motion consistency.
On top of Sandwich DiT, we introduce a two-in-one distillation pipeline that (i) aligns the student with a strong teacher model (Wan2.2) through offline cached distillation, and (ii) incorporates Distribution-Matching Distillation (DMD) and self-forcing to support few-step auto-regressive generation.

\subsection{Efficient Attention} \label{subsec:efficient attention}

To address the computation burden from the large latent space, we perform a systematic investigation on mobile-friendly efficient attention mechanisms. 
There are two major directions to reduce attention complexity, 
(i) reduce the quadratic complexity of self-attention by swapping to linear attention, 
and (ii) introduce token sparsity.

\subsubsection{LinConv Hybrid Attention~(LCHA)} \label{subsec:attnI}

Unlike conventional Softmax attention, Linear attention \cite{katharopoulos2020transformers} approximates the similarity function with a non-negative kernel which applies to queries and keys separately, which can be expressed as:
\[
y_i \;=\; \frac{\sum_{j=1}^{L} \phi(q_i)^{\!\top}\phi(k_j)\, v_j}{\sum_{j=1}^{L} \phi(q_i)^{\!\top}\phi(k_j)}
\quad\text{with }\phi(\cdot)\ge 0,
\]
where $L$ is the sequence length, $q_i,k_j$ are the query and key vectors at positions $i$ and $j$, and $v_j$ is the corresponding value.
According to associativity, we can then compute $\sum_{j=1}^{L} \phi(k_j) \cdot v_j^{\!\top}$ first, such that computation is linear to the number of tokens. 
But as discussed in \cite{qin2022cosformer,han2023flatten}, linear attention shows inferior modeling capability compared to full attention, and is more global but coarse in local information modeling. 
To address this, we propose a hybrid attention module that achieves linear computational complexity but captures both global and local dependency well.
As in \cref{fig:framework}, the module consists of two branches: 

\noindent \textbf{Linear Attention Path.} 
We introduce an optimized linear attention module.
Specifically, instead of using ReLU as the similarity kernel \cite{xie2024sana,chen2025sana}, we use a learnable positive kernel:
\begin{equation}
\begin{aligned}
&K(q,k) = \phi(q)^\top \phi(k),\\ 
&\phi(x) = \operatorname{softplus}(W x + b).
\end{aligned}
\end{equation}
This choice guarantees \(\phi(x)>0\), and the softplus preserves information for negative inputs and maintains non-vanishing gradients. 
We learn affine parameters \((W,b)\) end-to-end, allowing the feature map to adapt dynamically to distributional variation to improve the quality. 
Different from \cite{SANA}, we apply QK normalization \cite{henry2020query} and 3D RoPE embedding \cite{su2024roformer}, which were not used in their design.

\noindent \textbf{Local Conv Path.}
We augment the linear attention building block with a convolution path to enrich local detail modeling. 
Specifically, we employ a depth-wise 3D convolution followed by a linear channel mixing layer as a parallel branch. 
To support streaming generation, we apply temporal causal padding to the depth-wise 3D convolution.
According to our benchmark, dense 3D convolution will lead to OOM for mobile deployment. 

We use a learnable gate $\alpha$ (denoted as FusionGate) to mix the output of the two branches.
We include extensive ablation experiments in \cref{subsec:ablate efficient attention} to validate the effectiveness of the proposed LinConv Hybrid Attention (LCHA).

\subsubsection{Stride Self-Attention~(SSA)} \label{subsec:attnII}
Another route to efficient attention is token sparsity. 
From the scope of mobile support, only structured sparsity is of-interest in our case, while random or dynamic sparsity (sliding window) requires dedicated compilation, which is not yet supported. 
In this work, we investigate uniform KV sparsity (KV compression) and uniform QKV sparsity (stride self-attention).  
Among them, KV compression is a widely adopted strategy to reduce the cost of attention matrix multiplication while keeping the same output tokens   \cite{liu2018generating,wang2021pyramid}.
On the other hand, we can uniformly downsample QKV with certain strides, which requires an upsampling to restore the feature resolution. 
Note that empirically, using strided attention for all DiT blocks is equivalent to a low-resolution DiT, \ie, DiT on a higher compression latent space \cite{LTX-video,wu2025tamingdiffusiontransformerefficient}.

In our ablation study at \cref{subsec:ablate efficient attention}, we show that KV compression is inefficient, while relying solely on low-resolution SSA significantly degrades generation quality. Therefore, we employ the hybrid design that combines SSA with LCHA.
The two types of attention modules feature an elegant speed/quality trade-off, and provide a complementary design space that enables effective architecture search under mobile compute budgets.

\subsection{Sandwich Diffusion Transformer} \label{subsec:Search algorithm}

With the LCHA and SSA building blocks, we propose an automatic distribution search algorithm to form the final architecture. 
Compared to simple heuristics such as HDiT \cite{HourglassDiffusion}, which places low-resolution blocks in the middle to form a U-connected architecture, our automatically searched architecture distributes LCHA and SSA in an interleaved manner, as in \cref{fig:framework}. Our final model, Sandwich DiT, achieves higher performance compared to the HDiT architecture, which can be seen in \cref{subsec: ablation search}.

\subsubsection{Budget-aware Block Allocation.}
We first determine the counts of LCHA and SSA blocks via a budget-aware dynamic programming that satisfies the target latency and memory constraints.
Let $t\in\{\texttt{LCHA}, \texttt{SSA}\}$ denote the block type, where each transformer block employs the corresponding proposed attention.
Each type of block is associated with a latency $\ell_{t}$ and memory consumption $m_{t}$.
We choose non-negative integers $N_{t}$ such that satisfies with capacity $N_{\texttt{LCHA}}+N_{\texttt{SSA}}= K$, where $K$ is the total number of blocks, subject to a $\sum_{t} \ell_{t} N_{t} \le L_{\max}$,
and 
a peak-memory budget $\sum_{t} m_{t} N_{t} \le M_{\max}$.

To fully utilize the compute budget of the target device, we select the $(N_{\texttt{LCHA}}, N_{\texttt{SSA}})$ pair whose resulting $(L, M)$ is closest to $(L_{\max}, M_{\max})$.




\subsubsection{Attention Search Algorithm.}
Once the block counts of each type are fixed, we search for the optimal distribution of LCHA and SSA blocks to construct the final architecture.

\noindent \textbf{Problem Formulation.} We partition the search space ${\mathcal{G}}$ into $M$ groups $\{\mathcal{G}_n\}_{n=1}^M$, each contains $k$ blocks with LCHA or SSA.
At group $n$, a binary mask $m_n\!\in\!\{0,1\}$ routes computation to the LCHA ($m_n{=}1$) or SSA ($m_n{=}0$).
We maintain two feature streams, $y_L^{n}$ and $y_S^{n}$, and a buffer $S^{n}$ that caches the residual feature for long-skip connection.
Let $f^{n}_L(\cdot)$ and $f^{n}_S(\cdot)$ denote the $n$-th group of modules for the two branches, respectively.
Our goal is to learn $m_n$ that decides between $f^{n}_L(\cdot)$ or $f^{n}_S(\cdot)$ for each $n\in\{1,\dots,M\}$, yielding the optimal composition across groups.


\noindent \textbf{Branch-Switch Triggers.}
Since the two attention branches operate at different spatial resolutions, cross-branch routing at stage $n$ applies upsampling $\mathrm{up}_n(\cdot)$ and downsampling $\mathrm{down}_n(\cdot)$ as needed. We gate this choice with binary triggers $u_n$ and $d_n$ like following:
\begin{equation}
\begin{aligned}
u_n &= \max\{m_n - m_{n-1},\,0\},\\
d_n &= \max\{m_{n-1} - m_n,\,0\}.\\
\end{aligned}
\end{equation}
Thus, $u_n=1$ if routing switches from $y_S^{n-1}$ to $y_L^{n}$, and $d_n=1$ if it switches from $y_L^{n-1}$ to $y_S^{n}$, otherwise $u_n=d_n=0$.

\noindent \textbf{Per-group Differentiable Updating.}
As shown in \cref{fig:framework}, we first compute the switching results of both branches based on our definition:
\begin{equation}
\resizebox{0.95\linewidth}{!}{$
\begin{aligned}
\tilde y_L^{n} &= f_L^{n}\!\Big((1-u_n)\,y_L^{n-1}
  + u_n\bigl(\operatorname{up}(y_S^{n-1}) + S^{n-1}\bigr)\Big),\\
\tilde y_S^{n} &= f_{S}^{n}\!\Big((1-d_n)\,y_S^{n-1}
  + d_n\,\operatorname{down}\bigl(y_L^{n-1}\bigr)\Big)
\end{aligned}
$}
\end{equation}
where $\tilde y_L^{n}$ and $\tilde y_S^{n}$ are the switching results.
Besides, we update $S^{n}$ only at LCHA $\!\to\!$ SSA, which can be formulated as following:
\begin{equation}
S^{n} \;=\; d_n\,y_L^{n-1} \;+\; (1-d_n)\,S^{n-1}. 
\end{equation}
Finally, $m_n$ determines whether each stream is updated. we apply the Gumbel-Softmax function \cite{jang2016categorical} combined with the Straight-Through Estimator for the $m_n$ updating:
\begin{equation}
\begin{aligned}
y_L^{n} &= m_n\,\tilde y_L^{n} \;+\; (1-m_n)\,y_L^{n-1}, \\
y_S^{n} &= (1-m_n)\,\tilde y_S^{n} \;+\; m_n\,y_S^{n-1}.
\end{aligned}
\end{equation}
After the last group $M$, the output feature $\hat y$ is:
\begin{equation}
\hat y = m_M\cdot y_L^{(M)} + \left(1-m_M\right)\cdot\mathrm{up}\left(y_S^{(M)}\right).
\end{equation}

The details of our efficient search space and training design are in the \emph{supplementary materials}.


\subsection{2-in-1 Distillation} \label{sec: distill}

\noindent \textbf{Preliminaries.}
Our base model training follows the Rectified Flow ~\cite{wang2024rectified} objective. 
The forward path is a straight line from the data distribution $x_0$ to standard Gaussian noise $\epsilon \sim \mathcal{N}(0, I)$, the noisy sample $x_t$ at timestep $t\in[0,1]$ can be expressed as:
\begin{equation}
\setlength{\abovedisplayskip}{2pt} 
\setlength{\belowdisplayskip}{2pt}
x_t = (1 - t)\, x_0 + t\,\epsilon
\end{equation}
Along this path, the target velocity is $\epsilon-x_0$. We train our model $v_\theta$ to predict this velocity from $(x_t,t)$ via:
\begin{equation}
\setlength{\abovedisplayskip}{2pt} 
\setlength{\belowdisplayskip}{2pt}
\mathcal{L}_{\text{fm}}
=\mathbb{E}_{\epsilon\sim\mathcal{N}(0,I),\,t}\!\left[\left\|(\epsilon-x_0)-v_\theta(x_t,t)\right\|_2^2\right].
\end{equation}

%
\noindent \textbf{Offline Cached Knowledge Distillation.} 
Inspired by~\cite{wu2025tamingdiffusiontransformerefficient,fang2024tinyfusion}, we employ knowledge distillation as a final refinement stage to further improve \methodAbbr~after the architecture search and pretraining.
To enable effective knowledge transfer, we align the student and the teacher model by training both within an identical VAE latent space, and optimize the student to match the teacher’s outputs under identical conditioning and noise levels. 
Given that high quality video diffusion models like Huanyuan Video~\cite{HunyuanVideo}~(13B parameters) are impractically large and slow, requiring dozens of seconds for a single forward pass, an on-the-fly distillation approach is infeasible.

To address this issue, we introduce \emph{Offline Cached Knowledge Distillation}, a protocol that decouples teacher inference from student training by using precomputed, cached teacher outputs.
The cached data significantly improves training throughput and allows for larger batch sizes, which benefit convergence of the student model.
The proposed distillation pipeline consists of two stages:
\begin{enumerate*}[label=(\roman*)]
    \item Precompute and cache the noisy latents of real data, text embeddings, timestep and teacher model's output predictions.
    \item Perform distillation solely using the cached data, eliminating the need of real data, the text encoder, or teacher model during this stage. 
\end{enumerate*}
The distillation loss can be formulated as:
\begin{equation}
\setlength{\abovedisplayskip}{2pt} 
\setlength{\belowdisplayskip}{2pt}
\mathcal{L}_{\mathrm{KD}}
= \mathbb{E}\!\left[ \left\| u_{\theta_s}(x_t, t, c_\text{text}) - u_{\theta_t}(x_t, t, c_\text{text}) \right\|_2^2 \right],
\label{equ:kd}
\end{equation}
where $u_{\theta_s}$ is the student model, $x_t$ is noisy latent of timestep $t$, $ c_\text{text}$ is corresponding text embeddings, and $u_{\theta_t}(x_t, t, c_\text{text})$ is cached teacher output under the identical input. 
Specifically, we adopt Wan2.2-14B~\cite{wan2025wan} as the teacher model due to its superior visual fidelity.

\noindent \textbf{Distillation for Streaming.} \label{subsec:step distill}
Prior works such as APT2~\cite{lin2025autoregressiveadversarialposttrainingrealtime} and Self-Forcing~\cite{huang2025selfforcing}, address the training inference mismatch caused by teacher-forcing when generating video auto-regressively.
Following these approaches, we adopt distribution matching distillation~(DMD) to adapt the knowledge-distilled \methodAbbr~for auto-regressive~(chunk-level) video generation.
Before self-forcing, we collect the ODE trajectory of \methodAbbr~and perform teacher-forcing fine-tuning on bidirectional \methodAbbr~to obtain well-initialized generator weights.
During self-forcing DMD training, the real-score model is initialized from the same teacher as knowledge distillation~(\ie Wan2.2-14B), while the fake-score model is initialized from the knowledge-distilled \methodAbbr.

We also explore adversarial fine-tuning on top of the self-forcing DMD training.
Details are discussed in the \emph{supplementary materials}.

\noindent \textbf{Streaming Inference on Mobile.}
Thanks to the state inheritance, the causal inference in the Linear Attention and Causal Conv3D layers of the LCHA block requires only a fixed-size cache. 
In contrast, although the SSA block reduces the sequence length via a downsampling layer, its KV-cache still grows with the number of generated frames, leading to substantial memory overhead during mobile deployment.
To alleviate this issue, we apply window attention to the KV-cache, which both accelerates the inference speed and mitigates memory accumulation.

With DMD-based self-forcing fine-tuning, our model achieves efficient auto-regressive inference with fewer than 4-step per frame-chunk, enabling on-device, streaming video generation. 
Additional implementation and analysis details are available in the \emph{supplementary materials}.

\section{Experiments}

\begin{figure*}[t]
  \centering
  \includegraphics[width=1.\linewidth]{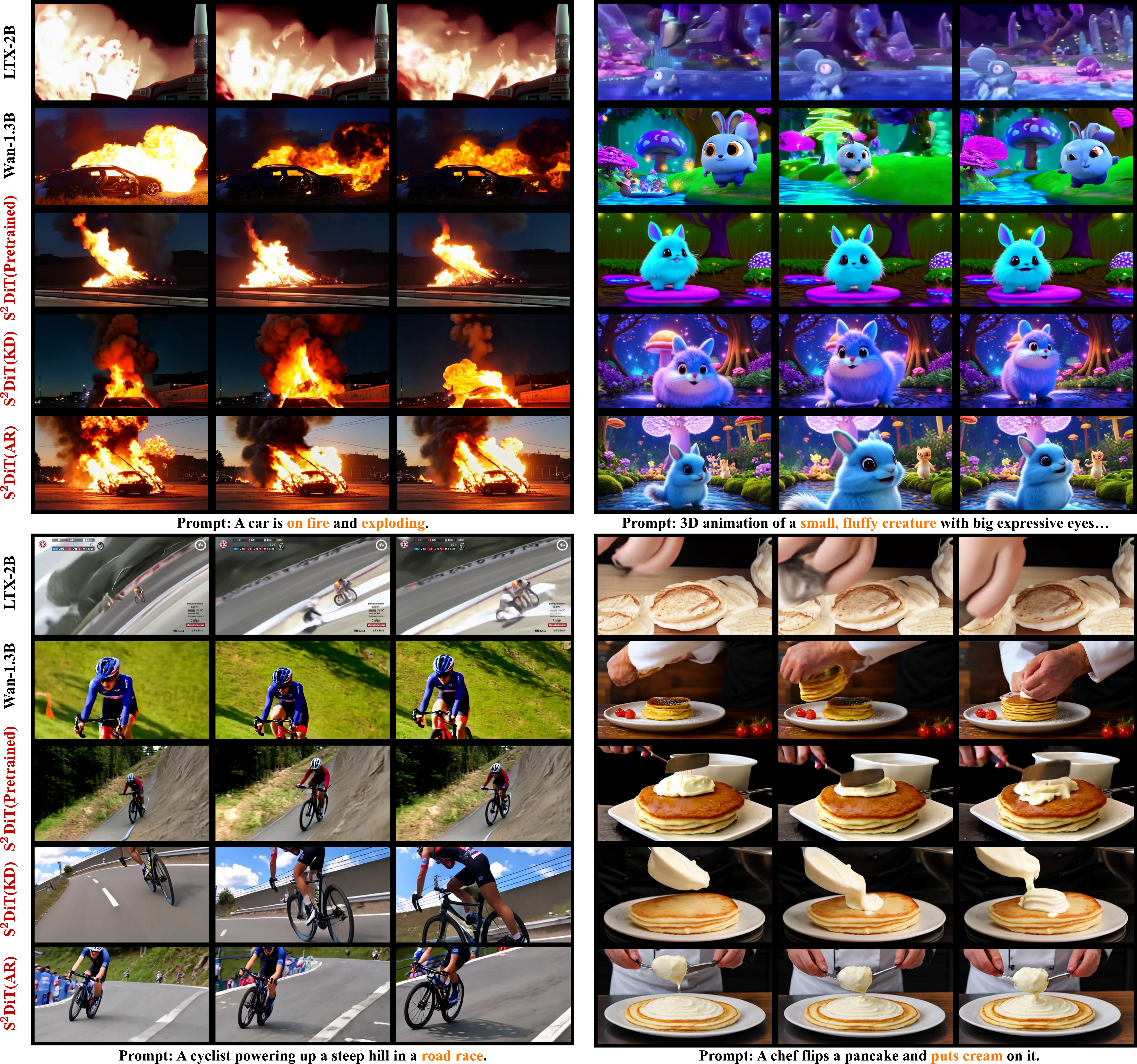}
  \caption{\textbf{Visual comparisons.} For Wan-1.3B \cite{wan2025wan} and LTX-2B 
  \cite{LTX-video}, videos are generated using their official default inference resolutions with the same prompts.}
  \label{fig:video}
\end{figure*}

\subsection{Setup}

\noindent \textbf{Data.} Following \cite{Panda70M,Allegro,wan2025wan}, we collect image and video data from both public and internal sources. After applying safety filters and quality filters (\eg, aesthetics, motion score), we arrive at a total of 300M images and 50M videos. 

\noindent \textbf{Model Pipeline.}
To achieve best generation quality, we employ the SOTA Wan Autoencoder with $4\times 8\times 8$ compression ratio. 
We further create an efficient decoder for mobile deployment, as detailed in \emph{supplementary material}. 
Similar to previous work \cite{SANA}, we employ CLIP-ViT-L \cite{radford2021learning} and Gemma3-4B-it \cite{team2025gemma} as the text encoder. CLIP serves as the efficient mobile model, while Gemma3 provides strong contextual information with prompt augmentation on-the-fly. 
We adopt the UniPC sampler~\cite{zhao2023unipc} following Wan2.1~\cite{wan2025wan}.

\noindent \textbf{Training Details.} 
Following progressive training \cite{wan2025wan}, our model is first pretrained on low resolution data for faster convergence, then scaled up to high resolution pretraining and knowledge distillation. 
We train 250K iterations at $256\times 144$ stage, and 50K iterations at $512\times288$ stage. 
Training is conducted on 256 NVIDIA A100 (80GB) GPUs using the AdamW optimizer with a learning rate of $1\times10^{-4}$ and $\beta=[0.9, 0.999]$.

\noindent \textbf{Evaluation.}
We compare Vbench score with popular video diffusion models.
For ablation studies, we report the evaluation loss with 50 sampling steps, CLIP score, FID, FVD.
We also include a user study in the \emph{supplementary material}. 

\noindent \textbf{Deployment.}
We use CoreMLTools \cite{coreml2021} to deploy the model to iPhone 16 Pro Max for speed benchmark and demonstration. We provide details in \emph{supplementary material}.

\begingroup
\setlength{\textfloatsep}{0pt}
\setlength{\intextsep}{0pt}
\begin{table*}[t]
\small
\centering
\resizebox{\linewidth}{!}{
\renewcommand\arraystretch{1}
\begin{tabular}{l|c|ccc|cccccc}
\toprule
Model & Params (B) & Total & Quality & Semantic & Flickering & Aesthetics & Imaging & Obj. Class & Scene & Consistency \\
\midrule
Hunyuan       & 13 & 83.24 & 85.09 & 75.82 & 99.44 & 60.36 & 67.56 & 86.10 & 53.88 & 26.44 \\
Open-Sora-2.0 & 11 & 84.34 & 85.40 & 80.12 & 99.40 & 64.39 & 65.66 & 94.50 & 52.71 & 27.50 \\
Open-Sora-1.2 & 1.2 & 79.76 & 81.35 & 73.39 & 99.53 & 56.85 & 63.34 & 82.22 & 42.44 & 26.85 \\
Wan2.1        & 14 & 84.70 & 85.64 & 80.95 & 99.53 & 61.53 & 67.28 & 94.24 & 53.67 & 27.44 \\
Wan2.1        & 1.3  & 83.31 & 85.23 & 75.65 & 99.55 & 65.46 & 67.01 & 88.81 & 41.96 & 25.50 \\
CogVideoX1.5  & 5  & 82.01 & 82.72 & 79.17 & 98.53 & 62.07 & 65.34 & 83.42 & 53.28 & 27.42 \\
CogVideoX     & 5  & 81.91 & 83.05 & 77.33 & 78.97 & 61.88 & 63.33 & 85.07 & 51.96 & 27.65 \\
LTX-Video     & 1.8  & 80.00 & 82.30 & 70.79 & 99.34 & 59.81 & 60.28 & 83.45 & 51.07 & 25.19 \\
SnapGen-V-DiT   & 0.9  & 81.45 & 83.12 & 74.76 & 98.11 & 64.16 & 63.41 & 92.26 & 51.06 & 25.51 \\ 
\midrule
\methodAbbr-Pretrained & 1.8  & 82.40 & 84.41 & 74.46 & 98.52 & 64.97 & 68.49 & 91.74 & 44.35 & 24.94 \\ 
\methodAbbr-KD &  1.8 &  83.62    &  86.13     &   73.58    &   99.56    &   65.26    &    69.05   &   91.76    &    48.37   &  25.35 \\
\methodAbbr-AR$\dagger$ &  1.8 &    83.26   &   85.63    &   73.79    &    98.20   &  65.64     &   70.57    &   89.49    &  49.19     & 24.75 \\

\bottomrule
\end{tabular}
}
\caption{\textbf{VBench~\citep{huang2023vbench} comparisons.} Scores for open-source models are collected from the~\href{https://huggingface.co/spaces/Vchitect/VBench_Leaderboard}{VBench Leaderboard}. 
$\dagger$ Tested with mobile deployment resolution ($512\times288$), other \methodAbbr~results are reported in higher resolution ($480\times832$) to better compare with baselines. 
}
\label{table: vbench comparison}
\end{table*}
\endgroup

\subsection{Qualitative Results}

We conduct a visual comparison of video samples generated by LTX-Video, Wan2.1-1.3B, \methodAbbr~(Pretrained), \methodAbbr~(Knowledge-Distilled), and \methodAbbr~(Auto-Regressive) in~\cref{fig:video}.
As illustrated, \methodAbbr~variants deliver superior video quality in terms of text-video alignment, photorealistic rendering, and smooth object motions compared to LTX-Video with heavily compressed latent space, while being comparable to Wan2.1-1.3B.
With knowledge distillation, \methodAbbr-(KD) produces vivid videos that surpass the quality of Wan2.1-1.3B.
The auto-regressive~(AR) variant achieves comparable visual fidelity to \methodAbbr-(KD), while requiring fewer sample steps and supporting efficient on-device streaming generation.
More visualizations are included in the \emph{supplementary material}.



\subsection{Quantitative Results}

We compare against recent open-source SOTA models, including LTX-Video, CogVideoX, Open-Sora, Wan2.1, and Hunyuan. Despite using only 1.8B parameters, \methodAbbr-(KD) attains a total VBench of 83.62, on par with Hunyuan-13B and Open-Sora-2.0, and close to Wan2.1-14B.
Our models also achieve strong Quality scores.
For fast inference, the \methodAbbr-(AR) reaches 83.26 while maintaining competitive fidelity, making it suitable for mobile deployment.
Besides, by comparing the three versions, \methodAbbr-(KD) and \methodAbbr-(AR) get better results than \methodAbbr-(Pretrained), which verifies the benefits of our proposed 2-in-1 Distillation in \cref{sec: distill}.

\subsection{Ablation Study} 


\subsubsection{Impact of Efficient Attention Design.} \label{subsec:ablate efficient attention}
We present an extensive ablation study that validates our efficient-attention design in \cref{tab:efftattn_ablation_abs_delta}. For this purpose, we randomly select $2,000$ videos from the OpenVid dataset \cite{nan2024openvid} that are disjoint from our training data.
The subset is utilized for the below ablations.
All variants are pre-trained under a matched compute budget and training schedule to ensure a fair comparison.

\noindent \textbf{Flat-attention variants.} \label{subsec:ablate single-stage}
We begin by assessing the variants in a single-stage setting as shown in the \emph{top block} of \cref{tab:efftattn_ablation_abs_delta}: 
(i) Full Attention: We set all blocks to standard self-attention. As shown in \cref{tab:efftattn_ablation_abs_delta}, this variant obtains the highest quality but is the most computationally expensive, owing to its quadratic complexity and full token budget. It can be seen as the upper-bound of our design, since our method serves as an efficient approximation.
\noindent (ii) LCHA-only: to assess the effect of the multi-stage design with mixing LCHA and SSA, we fix the attention to LCHA in all blocks.
\noindent (iii) SSA-only: similarly, we set all blocks to SSA for comparison, which reduces the token count by \(4\times\) and thus sacrifices the model capacity. As expected, this configuration yields the weakest quality and serves as the lower-bound for our model.

It shows that our result lies substantially closer to the upper-bound than the lower-bound across all metrics.
LCHA-only outperforms SSA-only across all metrics, indicating the importance of a high-resolution stage for generation. 
Besides, our two-stage model further surpasses LCHA-only on Eval loss, FID, and FVD, indicating that combining high-resolution modeling (LCHA) with low-resolution global context (SSA) yields the best results.

\noindent \textbf{Our Linear Attention vs. KV Compression.}
Beyond linear attention, key–value (KV) compression is also effective on mobile, as discussed in \cref{subsec:attnII}. 
Accordingly, we convert the linear attention path in LCHA with a KV compression variant, enabling a direct comparison to our proposed linear attention.
We evaluate both a one-stage model using KV compressed LCHA and a two-stage hourglass model combining KV compressed LCHA with SSA similar to our setting.
As shown in the \emph{middle block} of \cref{tab:efftattn_ablation_abs_delta}, the one-stage KV-Compression version is slightly better than LCHA-only.
However, based on the latency comparison, linear version get benefit compared to the kv-compression version.
For the two-stage model, linear version get a better result.

\noindent \textbf{Ablation of LCHA.}
Furthermore, we individually disable each proposed component to isolate its contribution in the LCHA design as demonstrated in the \emph{bottom block} of \cref{tab:efftattn_ablation_abs_delta}.
Removing either the linear path or the local path degrades quality. However, the local-only variant outperforms the linear-only variant, consistent with our hypothesis in \cref{subsec:attnI} that DiT-based models rely more heavily on local information.
We replace the PSoftplus mapping (linear + softplus) with a ReLU variant and observe worse FID/FVD, indicating that PSoftplus yields better perceptual quality.
We vary the head dimension $d_h$ of linear attention and observe that $d_h=256$ performs worse than $d_h=128$.
In addition, adding FusionGate yields a small but consistent improvement.
Our comprehensive experiments confirm that every component of LCHA is effective and synergistically contributes to the observed improvements.

\begingroup
\setlength{\textfloatsep}{0pt}
\setlength{\intextsep}{0pt}
\begin{table}[t]
  \centering
  \resizebox{\linewidth}{!}{
  \renewcommand\arraystretch{1}
\begin{tabular}{l|ccccc}
\toprule
Method                & Latency$\downarrow$ & Eval loss$\downarrow$ & CLIP$\uparrow$  & FID$\downarrow$    & FVD$\downarrow$     \\
\midrule
Full Attention        & OOM     & 0.252     & 0.281 & 32.035 & 317.130 \\
SSA-only              & 258     & 0.269     & 0.263 & 37.529 & 377.209 \\
LCHA-only             & 900     & 0.259     & 0.270 & 33.185 & 339.380 \\
\midrule
KV-Comp.-only         & 1348    & 0.255     & 0.275 & 32.609 & 333.473 \\
KV-Comp. + SSA         & 736     & 0.257     & 0.273 & 32.912 & 337.128 \\
\midrule
Local path-only       & 347     & 0.263     & 0.268 & 33.846 & 349.381 \\
Linear path-only      & 457     & 0.265     & 0.269 & 36.448 & 357.473 \\
PSoftplus$\rightarrow$ReLU      & 499     & 0.258     & 0.270 & 33.373 & 336.984 \\
LinAttn dim 128$\rightarrow$256 & 569     & 0.259     & 0.267 & 34.112 & 352.188 \\

\textit{w.o.} FusionGate       & 527     & 0.256     & 0.271 & 32.897 & 332.170 \\
\midrule
\cellcolor{LightCyan}Sandwich DiT                  & \cellcolor{LightCyan}531     & \cellcolor{LightCyan}0.256     & \cellcolor{LightCyan}0.272 & \cellcolor{LightCyan}32.209 & \cellcolor{LightCyan}330.978 \\
\bottomrule
\end{tabular}
  }
\caption{\textbf{Ablation of Our Efficient Attention Design .} Latency is benchmark on iPhone 16 Pro Max. \textbf{OOM} indicates out-of-memory.}
  \label{tab:efftattn_ablation_abs_delta}
\end{table}
\endgroup

\subsubsection{Impact of Sandwich Architecture.} \label{subsec: ablation search}
To assess the effectiveness of the proposed attention distribution search algorithm, we compare our Sandwich DiT with the Hourglass DiT under the same setting at $512\times288$ resolution.
The $512\times288$ models are obtained by fine-tuning from $256\times144$ models.
As illustrated in \cref{tab:model_res_cmp}, Sandwich DiT consistently outperforms the hourglass variant across metrics at both resolutions.

We further benchmark Full Attention (upper-bound) and SSA-only (lower-bound) at $512\times288$. 
Both hourglass and sandwich versions are closer to the upper-bound, consistent with \cref{subsec:ablate single-stage}, indicating that our design performs well at both low and high resolutions.

\begingroup
\setlength{\textfloatsep}{0pt}
\setlength{\intextsep}{0pt}
\begin{table}[t]
    \centering
    \footnotesize
    \resizebox{0.9\linewidth}{!}{
    \renewcommand\arraystretch{1}
    \begin{tabular}{l c |c c c c}
        \toprule
        Model & Res & $\mathcal{L}_{\text{eval}}$ $\downarrow$ & CLIP $\uparrow$ & FID $\downarrow$ & FVD $\downarrow$ \\
        \midrule
        Full Attention & $288\mathrm{p}$ & 0.180 & 0.294 & 26.495 & \ 130.825 \\
        SSA-only  & $288\mathrm{p}$ & 0.283 & 0.286 & 36.788 & 221.390 \\
        Hourglass & $288\mathrm{p}$ & 0.220 & 0.290 & 30.697 & 167.094 \\
        \cellcolor{LightCyan}Sandwich  & \cellcolor{LightCyan}$288\mathrm{p}$ & \cellcolor{LightCyan}0.209 & \cellcolor{LightCyan}0.291 & \ \cellcolor{LightCyan}29.565 & \cellcolor{LightCyan} 145.297 \\
        \bottomrule
    \end{tabular}
    }
    \caption{\textbf{Comparison of different architecture on $512 \times 288$ resolution.}}
    \label{tab:model_res_cmp}
\end{table}
\endgroup



\section{Conclusion}

We presented \methodAbbr, a Streaming Sandwich Diffusion Transformer that unifies architectural search, efficient training, and self-forcing inference for mobile video generation. By alternating hybrid linear-local and strided attention modules and optimizing their allocation through dynamic programming, \methodAbbr~achieves an exceptional balance between quality and latency. Our offline cached knowledge distillation pipeline enables compact students to inherit the fidelity of large-scale teachers at a fraction of the cost, combined with self-forcing, we further extend the model to real-time, auto-regressive video synthesis. Together, these designs push diffusion transformers beyond the server environment, demonstrating that high-quality streaming generation is achievable on mobile. 

\newpage
{
    \small
    \balance
    \bibliographystyle{ieeenat_fullname}
    \bibliography{main}

@String(CVPR= {IEEE Conf. Comput. Vis. Pattern Recog.})

@String(CVPR  = {CVPR})

@article{wan2025wan,
  title={Wan: Open and advanced large-scale video generative models},
  author={Wan, Team and Wang, Ang and Ai, Baole and Wen, Bin and Mao, Chaojie and Xie, Chen-Wei and Chen, Di and Yu, Feiwu and Zhao, Haiming and Yang, Jianxiao and others},
  journal={arXiv preprint arXiv:2503.20314},
  year={2025}
}

@article{lin2025autoregressiveadversarialposttrainingrealtime,
      title={Autoregressive Adversarial Post-Training for Real-Time Interactive Video Generation}, 
      author={Shanchuan Lin and Ceyuan Yang and Hao He and Jianwen Jiang and Yuxi Ren and Xin Xia and Yang Zhao and Xuefeng Xiao and Lu Jiang},
      year={2025},
      eprint={2506.09350},
      archivePrefix={arXiv},
      primaryClass={cs.CV},
      url={https://arxiv.org/abs/2506.09350}, 
}

@article{wu2025tamingdiffusiontransformerefficient,
      title={Taming Diffusion Transformer for Efficient Mobile Video Generation in Seconds}, 
      author={Yushu Wu and Yanyu Li and Anil Kag and Ivan Skorokhodov and Willi Menapace and Ke Ma and Arpit Sahni and Ju Hu and Aliaksandr Siarohin and Dhritiman Sagar and Yanzhi Wang and Sergey Tulyakov},
      year={2025},
      eprint={2507.13343},
      archivePrefix={arXiv},
      primaryClass={cs.CV},
      url={https://arxiv.org/abs/2507.13343}, 
}

@article{liu2018generating,
  title={Generating wikipedia by summarizing long sequences},
  author={Liu, Peter J and Saleh, Mohammad and Pot, Etienne and Goodrich, Ben and Sepassi, Ryan and Kaiser, Lukasz and Shazeer, Noam},
  journal={arXiv preprint arXiv:1801.10198},
  year={2018}
}

@inproceedings{wang2021pyramid,
  title={Pyramid vision transformer: A versatile backbone for dense prediction without convolutions},
  author={Wang, Wenhai and Xie, Enze and Li, Xiang and Fan, Deng-Ping and Song, Kaitao and Liang, Ding and Lu, Tong and Luo, Ping and Shao, Ling},
  booktitle={Proceedings of the IEEE/CVF international conference on computer vision},
  pages={568--578},
  year={2021}
}

@article{huang2025selfforcing,
  title={Self Forcing: Bridging the Train-Test Gap in Autoregressive Video Diffusion},
  author={Huang, Xun and Li, Zhengqi and He, Guande and Zhou, Mingyuan and Shechtman, Eli},
  journal={arXiv preprint arXiv:2506.08009},
  year={2025}
}

@article{dmd2,
 author = {Yin, Tianwei and Gharbi, Micha{\"e}l and Park, Taesung and Zhang, Richard and Shechtman, Eli and Durand, Fredo and Freeman, William T},
 journal = {ArXiv preprint},
 title = {Improved Distribution Matching Distillation for Fast Image Synthesis},
 url = {https://arxiv.org/abs/2405.14867},
 volume = {abs/2405.14867},
 year = {2024}
}

@inproceedings{han2023flatten,
  title={Flatten transformer: Vision transformer using focused linear attention},
  author={Han, Dongchen and Pan, Xuran and Han, Yizeng and Song, Shiji and Huang, Gao},
  booktitle={Proceedings of the IEEE/CVF international conference on computer vision},
  pages={5961--5971},
  year={2023}
}

@article{wang2024rectified,
  title={Rectified diffusion: Straightness is not your need in rectified flow},
  author={Wang, Fu-Yun and Yang, Ling and Huang, Zhaoyang and Wang, Mengdi and Li, Hongsheng},
  journal={arXiv preprint arXiv:2410.07303},
  year={2024}
}

@article{jang2016categorical,
  title={Categorical reparameterization with gumbel-softmax},
  author={Jang, Eric and Gu, Shixiang and Poole, Ben},
  journal={arXiv preprint arXiv:1611.01144},
  year={2016}
}

@article{hacohen2024ltx,
  title={Ltx-video: Realtime video latent diffusion},
  author={HaCohen, Yoav and Chiprut, Nisan and Brazowski, Benny and Shalem, Daniel and Moshe, Dudu and Richardson, Eitan and Levin, Eran and Shiran, Guy and Zabari, Nir and Gordon, Ori and others},
  journal={arXiv preprint arXiv:2501.00103},
  year={2024}
}

@article{zhao2023unipc,
  title={Unipc: A unified predictor-corrector framework for fast sampling of diffusion models},
  author={Zhao, Wenliang and Bai, Lujia and Rao, Yongming and Zhou, Jie and Lu, Jiwen},
  journal={Advances in Neural Information Processing Systems},
  volume={36},
  pages={49842--49869},
  year={2023}
}

@article{chen2025sana,
  title={Sana-video: Efficient video generation with block linear diffusion transformer},
  author={Chen, Junsong and Zhao, Yuyang and Yu, Jincheng and Chu, Ruihang and Chen, Junyu and Yang, Shuai and Wang, Xianbang and Pan, Yicheng and Zhou, Daquan and Ling, Huan and others},
  journal={arXiv preprint arXiv:2509.24695},
  year={2025}
}

@article{xie2024sana,
  title={Sana: Efficient high-resolution image synthesis with linear diffusion transformers},
  author={Xie, Enze and Chen, Junsong and Chen, Junyu and Cai, Han and Tang, Haotian and Lin, Yujun and Zhang, Zhekai and Li, Muyang and Zhu, Ligeng and Lu, Yao and others},
  journal={arXiv preprint arXiv:2410.10629},
  year={2024}
}

@inproceedings{henry2020query,
  title={Query-key normalization for transformers},
  author={Henry, Alex and Dachapally, Prudhvi Raj and Pawar, Shubham Shantaram and Chen, Yuxuan},
  booktitle={Findings of the Association for Computational Linguistics: EMNLP 2020},
  pages={4246--4253},
  year={2020}
}

@article{karnewar2025neodragon,
  title={Neodragon: Mobile Video Generation using Diffusion Transformer},
  author={Karnewar, Animesh and Korzhenkov, Denis and Lelekas, Ioannis and Karjauv, Adil and Fathima, Noor and Xiong, Hanwen and Vaidyanathan, Vancheeswaran and Zeng, Will and Esteves, Rafael and Singhal, Tushar and others},
  journal={arXiv preprint arXiv:2511.06055},
  year={2025}
}

@article{su2024roformer,
  title={Roformer: Enhanced transformer with rotary position embedding},
  author={Su, Jianlin and Ahmed, Murtadha and Lu, Yu and Pan, Shengfeng and Bo, Wen and Liu, Yunfeng},
  journal={Neurocomputing},
  volume={568},
  pages={127063},
  year={2024},
  publisher={Elsevier}
}

@article{xi2025sparse,
  title={Sparse videogen: Accelerating video diffusion transformers with spatial-temporal sparsity},
  author={Xi, Haocheng and Yang, Shuo and Zhao, Yilong and Xu, Chenfeng and Li, Muyang and Li, Xiuyu and Lin, Yujun and Cai, Han and Zhang, Jintao and Li, Dacheng and others},
  journal={arXiv preprint arXiv:2502.01776},
  year={2025}
}

@article{sun2024hunyuan,
  title={Hunyuan-large: An open-source moe model with 52 billion activated parameters by tencent},
  author={Sun, Xingwu and Chen, Yanfeng and Huang, Yiqing and Xie, Ruobing and Zhu, Jiaqi and Zhang, Kai and Li, Shuaipeng and Yang, Zhen and Han, Jonny and Shu, Xiaobo and others},
  journal={arXiv preprint arXiv:2411.02265},
  year={2024}
}

@article{hong2022cogvideo,
  title={Cogvideo: Large-scale pretraining for text-to-video generation via transformers},
  author={Hong, Wenyi and Ding, Ming and Zheng, Wendi and Liu, Xinghan and Tang, Jie},
  journal={arXiv preprint arXiv:2205.15868},
  year={2022}
}

@misc{opensora,
 author = {Zangwei Zheng and Xiangyu Peng and Tianji Yang and Chenhui Shen and Shenggui Li and Hongxin Liu and Yukun Zhou and Tianyi Li and Yang You},
 title = {Open-Sora: Democratizing Efficient Video Production for All},
 url = {https://github.com/hpcaitech/Open-Sora},
 year = {2024}
}

@misc{genmo2024mochi,
 author = {Genmo Team},
 title = {Mochi},
 year = {2024}
}

@article{polyak2024movie,
 author = {Polyak, Adam and Zohar, Amit and Brown, Andrew and Tjandra, Andros and Sinha, Animesh and Lee, Ann and Vyas, Apoorv and Shi, Bowen and Ma, Chih-Yao and Chuang, Ching-Yao and others},
 journal = {ArXiv preprint},
 title = {Movie gen: A cast of media foundation models},
 url = {https://arxiv.org/abs/2410.13720},
 volume = {abs/2410.13720},
 year = {2024}
}

@article{wang2024animatelcm,
 author = {Wang, Fu-Yun and Huang, Zhaoyang and Shi, Xiaoyu and Bian, Weikang and Song, Guanglu and Liu, Yu and Li, Hongsheng},
 journal = {ArXiv preprint},
 title = {Animatelcm: Accelerating the animation of personalized diffusion models and adapters with decoupled consistency learning},
 url = {https://arxiv.org/abs/2402.00769},
 volume = {abs/2402.00769},
 year = {2024}
}

@inproceedings{wu2023tune,
  title={Tune-a-video: One-shot tuning of image diffusion models for text-to-video generation},
  author={Wu, Jay Zhangjie and Ge, Yixiao and Wang, Xintao and Lei, Stan Weixian and Gu, Yuchao and Shi, Yufei and Hsu, Wynne and Shan, Ying and Qie, Xiaohu and Shou, Mike Zheng},
  booktitle={Proceedings of the IEEE/CVF international conference on computer vision},
  pages={7623--7633},
  year={2023}
}

@article{team2025gemma,
  title={Gemma 3 technical report},
  author={Team, Gemma and Kamath, Aishwarya and Ferret, Johan and Pathak, Shreya and Vieillard, Nino and Merhej, Ramona and Perrin, Sarah and Matejovicova, Tatiana and Ram{\'e}, Alexandre and Rivi{\`e}re, Morgane and others},
  journal={arXiv preprint arXiv:2503.19786},
  year={2025}
}

@inproceedings{radford2021learning,
  title={Learning transferable visual models from natural language supervision},
  author={Radford, Alec and Kim, Jong Wook and Hallacy, Chris and Ramesh, Aditya and Goh, Gabriel and Agarwal, Sandhini and Sastry, Girish and Askell, Amanda and Mishkin, Pamela and Clark, Jack and others},
  booktitle={International conference on machine learning},
  pages={8748--8763},
  year={2021},
  organization={PmLR}
}

@inproceedings{zhou2024upscale,
  title={Upscale-a-video: Temporal-consistent diffusion model for real-world video super-resolution},
  author={Zhou, Shangchen and Yang, Peiqing and Wang, Jianyi and Luo, Yihang and Loy, Chen Change},
  booktitle={Proceedings of the IEEE/CVF Conference on Computer Vision and Pattern Recognition},
  pages={2535--2545},
  year={2024}
}

@inproceedings{huang2023vbench,
 author = {Huang, Ziqi and He, Yinan and Yu, Jiashuo and Zhang, Fan and Si, Chenyang and Jiang, Yuming and Zhang, Yuanhan and Wu, Tianxing and Jin, Qingyang and Chanpaisit, Nattapol and Wang, Yaohui and Chen, Xinyuan and Wang, Limin and Lin, Dahua and Qiao, Yu and Liu, Ziwei},
 booktitle = {Proceedings of the IEEE/CVF Conference on Computer Vision and Pattern Recognition},
 title = {{VBench}: Comprehensive Benchmark Suite for Video Generative Models},
 year = {2024}
}

@article{nan2024openvid,
 author = {Nan, Kepan and Xie, Rui and Zhou, Penghao and Fan, Tiehan and Yang, Zhenheng and Chen, Zhijie and Li, Xiang and Yang, Jian and Tai, Ying},
 journal = {ArXiv preprint},
 title = {Openvid-1m: A large-scale high-quality dataset for text-to-video generation},
 url = {https://arxiv.org/abs/2407.02371},
 volume = {abs/2407.02371},
 year = {2024}
}

@misc{coreml2021,
 author = {{Apple Inc.}},
  title = {{Core ML Tools}},
  howpublished = {\url{https://coremltools.readme.io/}},
  year = {2024},
  note = {Version 8.0, accessed on September 7, 2025},
}

@misc{kling,
 author = {Kuaishou},
 howpublished = {\url{https://kling.kuaishou.com/en}},
 title = {Kling},
 year={2024}
}

@misc{sora,
 author = {OpenAI},
 howpublished = {\url{https://openai.com/index/video-generation-models-as-world-simulators/}},
 title = {Video generation models as world simulators},
 year={2023}
}

@article{wan2025wanopenadvancedlargescale,
      title={Wan: Open and Advanced Large-Scale Video Generative Models}, 
      author={Team Wan and Ang Wang and Baole Ai and Bin Wen and Chaojie Mao and Chen-Wei Xie and Di Chen and Feiwu Yu and Haiming Zhao and Jianxiao Yang and Jianyuan Zeng and Jiayu Wang and Jingfeng Zhang and Jingren Zhou and Jinkai Wang and Jixuan Chen and Kai Zhu and Kang Zhao and Keyu Yan and Lianghua Huang and Mengyang Feng and Ningyi Zhang and Pandeng Li and Pingyu Wu and Ruihang Chu and Ruili Feng and Shiwei Zhang and Siyang Sun and Tao Fang and Tianxing Wang and Tianyi Gui and Tingyu Weng and Tong Shen and Wei Lin and Wei Wang and Wei Wang and Wenmeng Zhou and Wente Wang and Wenting Shen and Wenyuan Yu and Xianzhong Shi and Xiaoming Huang and Xin Xu and Yan Kou and Yangyu Lv and Yifei Li and Yijing Liu and Yiming Wang and Yingya Zhang and Yitong Huang and Yong Li and You Wu and Yu Liu and Yulin Pan and Yun Zheng and Yuntao Hong and Yupeng Shi and Yutong Feng and Zeyinzi Jiang and Zhen Han and Zhi-Fan Wu and Ziyu Liu},
      year={2025},
      eprint={2503.20314},
      archivePrefix={arXiv},
      primaryClass={cs.CV},
      url={https://arxiv.org/abs/2503.20314}, 
}

@article{hacohen2024ltxvideorealtimevideolatent,
      title={LTX-Video: Realtime Video Latent Diffusion}, 
      author={Yoav HaCohen and Nisan Chiprut and Benny Brazowski and Daniel Shalem and Dudu Moshe and Eitan Richardson and Eran Levin and Guy Shiran and Nir Zabari and Ori Gordon and Poriya Panet and Sapir Weissbuch and Victor Kulikov and Yaki Bitterman and Zeev Melumian and Ofir Bibi},
      year={2024},
      eprint={2501.00103},
      archivePrefix={arXiv},
      primaryClass={cs.CV},
      url={https://arxiv.org/abs/2501.00103}, 
}

@inproceedings{wu2024snapgenvgeneratingfivesecondvideo,
      title={SnapGen-V: Generating a Five-Second Video within Five Seconds on a Mobile Device}, 
      author={Wu, Yushu and Zhang, Zhixing and Li, Yanyu and Xu, Yanwu and Kag, Anil and Sui, Yang and Coskun, Huseyin and Ma, Ke and Lebedev, Aleksei and Hu, Ju and others},
      booktitle={Proceedings of the Computer Vision and Pattern Recognition Conference},
      pages={2479--2490},
      year={2025}
}

@article{kim2025ondevicesoraenablingtrainingfree,
      title={On-device Sora: Enabling Training-Free Diffusion-based Text-to-Video Generation for Mobile Devices}, 
      author={Bosung Kim and Kyuhwan Lee and Isu Jeong and Jungmin Cheon and Yeojin Lee and Seulki Lee},
      year={2025},
      eprint={2502.04363},
      archivePrefix={arXiv},
      primaryClass={cs.CV},
      url={https://arxiv.org/abs/2502.04363}, 
}

@article{yahia2024mobilevideodiffusion,
      title={Mobile Video Diffusion}, 
      author={Haitam Ben Yahia and Denis Korzhenkov and Ioannis Lelekas and Amir Ghodrati and Amirhossein Habibian},
      year={2024},
      eprint={2412.07583},
      archivePrefix={arXiv},
      primaryClass={cs.CV},
      url={https://arxiv.org/abs/2412.07583}, 
}

@String(CVPR = {IEEE/CVF Conference on Computer Vision and Pattern Recognition (CVPR)})

@String(NeurIPS  = {Advances in Neural Information Processing Systems (NeurIPS)})

@article{qin2022cosformer,
  title={cosformer: Rethinking softmax in attention},
  author={Qin, Zhen and Sun, Weixuan and Deng, Hui and Li, Dongxu and Wei, Yunshen and Lv, Baohong and Yan, Junjie and Kong, Lingpeng and Zhong, Yiran},
  journal={arXiv preprint arXiv:2202.08791},
  year={2022}
}

@inproceedings{katharopoulos2020transformers,
  title={Transformers are rnns: Fast autoregressive transformers with linear attention},
  author={Katharopoulos, Angelos and Vyas, Apoorv and Pappas, Nikolaos and Fleuret, Fran{\c{c}}ois},
  booktitle={International conference on machine learning},
  pages={5156--5165},
  year={2020},
  organization={PMLR}
}

@inproceedings{FID,
  title={Gans trained by a two time-scale update rule converge to a local nash equilibrium},
  author={Heusel, Martin and Ramsauer, Hubert and Unterthiner, Thomas and Nessler, Bernhard and Hochreiter, Sepp},
  booktitle=NeurIPS,
  year={2017}
}

@article{ImagenVideo,
  title={Imagen video: High definition video generation with diffusion models},
  author={Ho, Jonathan and Chan, William and Saharia, Chitwan and Whang, Jay and Gao, Ruiqi and Gritsenko, Alexey and Kingma, Diederik P and Poole, Ben and Norouzi, Mohammad and Fleet, David J and others},
  journal={arXiv preprint arXiv:2210.02303},
  year={2022}
}

@article{AnimateDiff,
  title={Animatediff: Animate your personalized text-to-image diffusion models without specific tuning},
  author={Guo, Yuwei and Yang, Ceyuan and Rao, Anyi and Wang, Yaohui and Qiao, Yu and Lin, Dahua and Dai, Bo},
  journal={arXiv preprint arXiv:2307.04725},
  year={2023}
}

@inproceedings{zhang2024sfv,
    title={{SF}-V: Single Forward Video Generation Model},
    author={Zhixing Zhang and Yanyu Li and Yushu Wu and yanwu xu and Anil Kag and Ivan Skorokhodov and Willi Menapace and Aliaksandr Siarohin and Junli Cao and Dimitris N. Metaxas and Sergey Tulyakov and Jian Ren},
    booktitle={The Thirty-eighth Annual Conference on Neural Information Processing Systems},
    year={2024},
    url={https://openreview.net/forum?id=PVgAeMm3MW}
  }

@article{SnapVideo,
  title={Snap Video: Scaled Spatiotemporal Transformers for Text-to-Video Synthesis},
  author={Menapace, Willi and Siarohin, Aliaksandr and Skorokhodov, Ivan and Deyneka, Ekaterina and Chen, Tsai-Shien and Kag, Anil and Fang, Yuwei and Stoliar, Aleksei and Ricci, Elisa and Ren, Jian and others},
  journal={arXiv preprint arXiv:2402.14797},
  year={2024}
}

@article{HourglassDiffusion,
  title={Scalable High-Resolution Pixel-Space Image Synthesis with Hourglass Diffusion Transformers},
  author={Crowson, Katherine and Baumann, Stefan Andreas and Birch, Alex and Abraham, Tanishq Mathew and Kaplan, Daniel Z and Shippole, Enrico},
  journal={arXiv preprint arXiv:2401.11605},
  year={2024}
}

@article{SVD,
  title={Stable video diffusion: Scaling latent video diffusion models to large datasets},
  author={Blattmann, Andreas and Dockhorn, Tim and Kulal, Sumith and Mendelevitch, Daniel and Kilian, Maciej and Lorenz, Dominik and Levi, Yam and English, Zion and Voleti, Vikram and Letts, Adam and others},
  journal={arXiv preprint arXiv:2311.15127},
  year={2023}
}

@article{HunyuanVideo,
  title={HunyuanVideo: A Systematic Framework For Large Video Generative Models},
  author={Kong, Weijie and Tian, Qi and Zhang, Zijian and Min, Rox and Dai, Zuozhuo and Zhou, Jin and Xiong, Jiangfeng and Li, Xin and Wu, Bo and Zhang, Jianwei and others},
  journal={arXiv preprint arXiv:2412.03603},
  year={2024}
}

@article{MovieGen,
  title={Movie gen: A cast of media foundation models},
  author={Polyak, Adam and Zohar, Amit and Brown, Andrew and Tjandra, Andros and Sinha, Animesh and Lee, Ann and Vyas, Apoorv and Shi, Bowen and Ma, Chih-Yao and Chuang, Ching-Yao and others},
  journal={arXiv preprint arXiv:2410.13720},
  year={2024}
}

@article{LTX-video,
  title={Ltx-video: Realtime video latent diffusion},
  author={HaCohen, Yoav and Chiprut, Nisan and Brazowski, Benny and Shalem, Daniel and Moshe, Dudu and Richardson, Eitan and Levin, Eran and Shiran, Guy and Zabari, Nir and Gordon, Ori and others},
  journal={arXiv preprint arXiv:2501.00103},
  year={2024}
}

@article{SANA,
  title={Sana: Efficient high-resolution image synthesis with linear diffusion transformers},
  author={Xie, Enze and Chen, Junsong and Chen, Junyu and Cai, Han and Tang, Haotian and Lin, Yujun and Zhang, Zhekai and Li, Muyang and Zhu, Ligeng and Lu, Yao and others},
  journal={arXiv preprint arXiv:2410.10629},
  year={2024}
}

@article{Allegro,
  title={Allegro: Open the black box of commercial-level video generation model},
  author={Zhou, Yuan and Wang, Qiuyue and Cai, Yuxuan and Yang, Huan},
  journal={arXiv preprint arXiv:2410.15458},
  year={2024}
}

@article{opensoraplan,
  title={Open-sora plan: Open-source large video generation model},
  author={Lin, Bin and Ge, Yunyang and Cheng, Xinhua and Li, Zongjian and Zhu, Bin and Wang, Shaodong and He, Xianyi and Ye, Yang and Yuan, Shenghai and Chen, Liuhan and others},
  journal={arXiv preprint arXiv:2412.00131},
  year={2024}
}

@article{CogVideoX,
  title={Cogvideox: Text-to-video diffusion models with an expert transformer},
  author={Yang, Zhuoyi and Teng, Jiayan and Zheng, Wendi and Ding, Ming and Huang, Shiyu and Xu, Jiazheng and Yang, Yuanming and Hong, Wenyi and Zhang, Xiaohan and Feng, Guanyu and others},
  journal={arXiv preprint arXiv:2408.06072},
  year={2024}
}

@inproceedings{Panda70M,
  title     = {Panda-70M: Captioning 70M Videos with Multiple Cross-Modality Teachers},
  author    = {Chen, Tsai-Shien and Siarohin, Aliaksandr and Menapace, Willi and Deyneka, Ekaterina and Chao, Hsiang-wei and Jeon, Byung Eun and Fang, Yuwei and Lee, Hsin-Ying and Ren, Jian and Yang, Ming-Hsuan and Tulyakov, Sergey},
  booktitle = {Proceedings of the IEEE/CVF Conference on Computer Vision and Pattern Recognition},
  year      = {2024}
}

@article{StepVideoT2V,
  title={Step-Video-T2V Technical Report: The Practice, Challenges, and Future of Video Foundation Model},
  author={Ma, Guoqing and Huang, Haoyang and Yan, Kun and Chen, Liangyu and Duan, Nan and Yin, Shengming and Wan, Changyi and Ming, Ranchen and Song, Xiaoniu and Chen, Xing and others},
  journal={arXiv preprint arXiv:2502.10248},
  year={2025}
}

@article{fang2024tinyfusion,
  title={TinyFusion: Diffusion Transformers Learned Shallow},
  author={Fang, Gongfan and Li, Kunjun and Ma, Xinyin and Wang, Xinchao},
  journal={arXiv preprint arXiv:2412.01199},
  year={2024}
}

@inproceedings{yin2025causvid,
    title={From Slow Bidirectional to Fast Autoregressive Video Diffusion Models},
    author={Yin, Tianwei and Zhang, Qiang and Zhang, Richard and Freeman, William T and Durand, Fredo and Shechtman, Eli and Huang, Xun},
    booktitle={CVPR},
    year={2025}
}

@article{chen2025sanavideoefficientvideogeneration,
      title={SANA-Video: Efficient Video Generation with Block Linear Diffusion Transformer}, 
      author={Junsong Chen and Yuyang Zhao and Jincheng Yu and Ruihang Chu and Junyu Chen and Shuai Yang and Xianbang Wang and Yicheng Pan and Daquan Zhou and Huan Ling and Haozhe Liu and Hongwei Yi and Hao Zhang and Muyang Li and Yukang Chen and Han Cai and Sanja Fidler and Ping Luo and Song Han and Enze Xie},
      year={2025},
      eprint={2509.24695},
      archivePrefix={arXiv},
      primaryClass={cs.CV},
      url={https://arxiv.org/abs/2509.24695}, 
}
}
\newpage
\clearpage
\appendix
\setcounter{page}{1}
\maketitlesupplementary

\setcounter{section}{0}
\setcounter{figure}{0}
\setcounter{table}{0}
\renewcommand{\thefigure}{A\arabic{figure}}
\renewcommand{\thetable}{A\arabic{table}}


\section{Search algorithm}

We provide the detailed search algorithm as follows \cref{alg:sandwich}. Moreover, we adopt design improvements that dramatically reduce computation and yield higher training efficiency.

\noindent \textbf{Efficient Search Space Formulation.}
We reduce the search space in three ways: (i) We use group-wise rather than block-wise masking, yielding an exponential reduction in the search space. With two blocks per group, the space reduces from $2^{(2n)}$ to $2^{n}$. (ii) We set $m_1=m_M=1$ to make the spatial scale matches the VAE compression ratio, further reducing the space to $2^{(n-2)}$. (iii) Base on the dynamic programming, we fix the counts of groups assigned to LCHA ($k$) and SSA ($n-k-2$) in the search space, reducing the space from $2^{(n-2)}$ to $C_{n-2}^{k}$.
Overall, these choices reduce the space from $2^{(2n)}$ to $C_{n-2}^{k}$, yielding a substantially more efficient and well-structured search algorithm.

\begin{algorithm}[ht]
\small
\caption{Sandwich Architecture Search Algorithm}
\label{alg:sandwich}
\begin{algorithmic}[1]
\STATE \textbf{Input:} Input feature $y$, groups $M$, group modules $\{(f_L^{n}, f_S^{n})\}_{n=1}^{M}$, 
upsampler $\mathrm{up}_n(\cdot)$, downsampler $\mathrm{down}_n(\cdot)$
\STATE \textbf{Learnable:} Binary routers $\{m_n\}_{n=1}^{M}$ (trained via Gumbel-Softmax + STE)
\STATE \textbf{Output:} Output feature $\hat{y}$
\STATE Initialize $y_L^{0} \leftarrow y$, \ $y_S^{0} \leftarrow \mathrm{down}_1(y)$
\STATE Initialize $S^{0} \leftarrow 0$ 
\STATE Initialize $m_0 \leftarrow 1$ 
\vspace{2pt}
\FOR{$n = 1$ \TO $M$}

    \item[] \texttt{// Branch-Switch Triggers}
    
    \STATE $u_n \leftarrow \max\{m_n - m_{n-1}, \ 0\}$ 
    \STATE $d_n \leftarrow \max\{m_{n-1} - m_n, \ 0\}$ 

    \vspace{2pt}
    \item[] \texttt{// Prepare Inputs for Group $n$}

    \STATE $x_L^{n} \leftarrow (1-u_n)\,y_L^{n-1} \;+\; u_n \cdot \bigl(\mathrm{up}_n(y_S^{n-1}) + S^{n-1}\bigr)$
    \STATE $x_S^{n} \leftarrow (1-d_n)\,y_S^{n-1} \;+\; d_n \cdot \mathrm{down}_n(y_L^{n-1})$
    
    \vspace{2pt}
    \item[] \texttt{// Per-Group Forward}

    \STATE $\tilde y_L^{n} \leftarrow f_L^{n}(x_L^{n})$ 
    \STATE $\tilde y_S^{n} \leftarrow f_S^{n}(x_S^{n})$
    
    \vspace{2pt}
    \item[] \texttt{// Update Long-Skip Buffer} 

    \STATE $S^{n} \leftarrow d_n \cdot y_L^{n-1} \;+\; (1-d_n)\cdot S^{n-1}$
    
    \vspace{2pt}
    \item[] \texttt{ // Differentiable Routing via Gumbel-Softmax + STE}
   
    \STATE $y_L^{n} \leftarrow m_n \cdot \tilde y_L^{n} \;+\; (1-m_n)\cdot y_L^{n-1}$
    \STATE $y_S^{n} \leftarrow (1-m_n)\cdot \tilde y_S^{n} \;+\; m_n \cdot y_S^{n-1}$
\ENDFOR

\item[] \texttt{ // Final Merge}
\STATE $\hat y \leftarrow m_M \cdot y_L^{(M)} \;+\; (1-m_M)\cdot \mathrm{up}_M\!\bigl(y_S^{(M)}\bigr)$
\vspace{2pt}
\STATE \textbf{return} $\hat y$
\end{algorithmic}
\end{algorithm}

\noindent \textbf{Efficient training design.}
We replace costly pre-training with self-distillation to achieve efficient training, as expressed as:
\begin{equation}
\begin{split}
   \mathcal{L}_{\mathrm{sd}}
= \mathbb{E}\!\left[ \left\| \hat y - T_{\theta}(y_{L}^1) \right\|_2^2\right],
\end{split}
\label{equ:kd_app}
\end{equation}
where $\mathcal{L}_{\mathrm{sd}}$ is the self-distillation loss, $T_{\theta}(.)$ denotes the pretrained teacher model with all self-attention blocks.
For the student model, we initialize the LCHA and SSA  by inheriting parameters from $T_{\theta}(.)$ wherever compatible.
Since both LCHA and SSA are variants of self-attention, most parameters are transferable, yielding a more efficient training setup.

\section{Model Design}
\noindent \textbf{RoPE-3D.}
We follow RoPE~\cite{su2024roformer} to incorporate rotary position embeddings into the linear attention formulation, as shown in~\cref{equ:rope linear}.
As discussed in~\cite{su2024roformer}, RoPE injects position while keep norm unchanged.
Therefore, the RoPE transformation is applied only to the outputs of non-linear functions.
Meanwhile, the denominator remains unchanged to avoid potential division-by-zero issues.

\begin{equation}
\label{equ:rope linear}
\text{Attn}(\mathbf{Q}, \mathbf{K}, \mathbf{V})_m =
\frac{
\sum_{n=1}^{N} \left( \mathbf{R}_{\Theta,m}^d \phi(\mathbf{q}_m) \right)^{\top} 
\left( \mathbf{R}_{\Theta,n}^d \varphi(\mathbf{k}_n) \right) \mathbf{v}_n
}{
\sum_{n=1}^{N} \phi(\mathbf{q}_m)^{\top} \varphi(\mathbf{k}_n)
}.
\end{equation}

\noindent \textbf{Normalization.} We employ Layer Normalization in transformer blocks and also as QK norm. 
In mobile deployment we identify that RMS norm yields higher numeric error, as a result, we use LayerNorm with affine transformation instead. 
We use AdaLN as timestep encoding which follows common practices \cite{wan2025wan,LTX-video}. 

\section{More details for 2-in-1 Distillation}

\subsection{Details of Offline Cached Knowledge Distillation.}
After evaluating visual fidelity across several candidates, we adopt Wan2.2-14B~\cite{wan2025wan} as the teacher. A key challenge is Wan2.2’s Mixture-of-Experts design, with separate high-noise and low-noise experts, which makes on-the-fly distillation computationally prohibitive.
We provide the details of our distillation procedure here. 

We propose \emph{Offline Cached Knowledge Distillation}, an offline two-stage protocol:
(i) \textit{Cache stage}: for the teacher model, we precompute and cache text embedding $e_t$, the diffusion tuple of timestep, noise, and velocity of high noise expert $(t_h, n_h, v_h)$ and low noise expert $(t_l, n_l, v_l)$.
(ii) \textit{Distillation stage}: during distillation, the training only uses cached tuples and skips teacher forward passes, which substantially reduces both FLOPs and peak memory.
The process can be formally defined by:
\begin{equation}
\resizebox{0.885\linewidth}{!}{
   $\mathcal{L}_{\mathrm{KD}}
= \mathbb{E}\!\left[ w_l\left\| v_l - V_{\theta}(t_l, n_l, e_t) \right\|_2^2 + w_h\left\| v_h - V_{\theta}(t_h, n_h, e_h) \right\|_2^2 \right],$
}
\label{equ:kd_cache}
\end{equation}
where $V_{\theta}(.)$ indicates the predicted velocity of our model, $w_l$ and $w_h$ are the hyper-parameter to adjust the weight between the two experts.

We set $w_l=w_h=0.5$ in our experiments.
\subsection{Details of Self-Forcing Distillation.}
Self-Forcing~\cite{huang2025selfforcing} fine-tuning pipeline has shown promising results for 4-step auto-regressive generation.
However, its performance tends to degrade significantly when applied to fewer steps (\eg, 1-step or 2-step generation).
A common approach to address this issue is adversarial fine-tuning, which aims to enhance quality in low-step generation.
Nonetheless, adversarial fine-tuning often suffers from training instability due to the substantial gap between ``real" and ``fake" samples, since previous works~\cite{zhang2024sfv, wu2024snapgenvgeneratingfivesecondvideo, dmd2} typically use real-world data as the ``real" samples. 
To mitigate the misalignment, a more intuitive strategy is to adopt progressive adversarial fine-tuning, where samples generated with more sampling steps are treated as the ``real" samples instead of real-world data.
Following backward simulation~\cite{dmd2}, we denote $x_{0;T}$ and $x_{0;T^\prime}$ as the $x_0$ predictions obtained with $T$ and $T^\prime$ sampling steps, respectively, where $T < T^\prime$. In this setting, $x_{0;T}$ serves as the “fake” sample and $x_{0;T^\prime}$ as the “real” sample.
Inspired by R3GAN, we employ the RpGAN$+R_1+R_2$ formulation as the adversarial fine-tuning objective:
\begin{equation}
    \begin{split}
    & R_1(\psi)=\frac{\gamma}{2}\frac{1}{\epsilon}\mathbb{E}\left[\mathcal{D}(x_{t;T^\prime}+\epsilon) - \mathcal{D}(x_{t;T^\prime})\right],
    \\
    &
    R_2(\psi)=\frac{\gamma}{2}\frac{1}{\epsilon}\mathbb{E}\left[\mathcal{D}(x_{t;T}+\epsilon) - \mathcal{D}(x_{t;T})\right],
    \end{split}
\end{equation}
\begin{equation}
    \begin{split} 
    \mathcal{L^\mathcal{D}_{\text{adv}}}=&\mathbb{E}\left[ \text{softplus}\left(- (\mathcal{D}(x_{t;T^\prime}) - \mathcal{D}(x_{t;T}))\right) \right],
    \\
    \mathcal{L^\mathcal{G}_{\text{adv}}}=&\mathbb{E}\left[ \text{softplus}\left(\mathcal{D}(x_{t;T^\prime}) - \mathcal{D}(x_{t;T})\right) \right],
    \end{split}
\end{equation}
where the $x_{t;T^\prime}$ and $x_{t;T}$ are noisy latent that backward simulated with $T$ and $T^\prime$ sampling step added by $t$ noise-level.

\noindent \textbf{Training details.}
We adopt the AdamW optimizer for the generator and discriminator, using a learning rate of $1.0e-5$ for the generator and $2.0e-6$ for the discriminator with betas set to [0, 0.999].
An exponential moving average (EMA) with a decay rate of $0.99$ is also applied to the generator for improved training stability.

\begingroup
\setlength{\tabcolsep}{3pt} 
\begin{table}[!htb]
\centering
\small
\resizebox{\linewidth}{!}{
\begin{tabular}{l | c c c c c c}
\toprule

\bf{\# Steps}& \bf{DD}& \bf{OC} & \bf{AQ} & \bf{Quality} & \bf{Semantic} &  \bf{Total} \\
\midrule
$1$ & $55.09$ & $85.42$& $52.85$ & $82.80$ & $70.65$ & $80.37$ \\
$2$ & $57.50$ & $91.25$& $64.04$ & $85.08$ & $73.15$ & $82.69$ \\
$4$ & $59.17$ & $89.49$& $65.64$ & $85.26$ & $73.79$ & $83.26$ \\
\bottomrule
\end{tabular}
}
\caption{\textbf{Analysis of the number of inference steps.} We measure VBench~\cite{huang2023vbench} score with different numbers of inference steps. In the results, ``DD", ``OC", and ``AQ" denote the dynamic degree, object class, and aesthetic quality scores, respectively.}
\label{table: ab inference}
\end{table}
\endgroup
 
\noindent \textbf{Evaluation on inference-step.} We provide the performance of different inference step in \cref{table: ab inference}. A single step already delivers surprisingly strong results with 80.37 total score, indicating that most of the global structure and appearance can be formed in one-step. 
Besides, the two-step setting provides the best speed–quality trade-off and the most stable behavior across metrics. 
Finally, increasing to four steps further improves accuracy, reaching the highest score (83.26) with acceptable latency.
%


\section{Details of Mobile Deployment}

\subsection{Efficient Decoder}

We freeze the VAE encoder of Wan2.1 and train an efficient decoder that decodes in real-time on mobile, as shown in (\cref{tab:efficientdecoder}). 
We build the efficient decoder with narrow 3D convolutions, group normalizations, and use Hardswish as the activation function. 
We encode video data with Wan2.1 encoder and obtain the latents, and the efficient decoder is trained to reconstruct the video with L1 loss, Perceptual loss and GAN loss. Our efficient decoder is $4\times$ smaller in size and can decode beyond real time on mobile, while still deliver on-par reconstruction quality. 

\begin{table}[ht]
\small
\centering
\begin{tabular}{cccccc}
\toprule
Decoder & Params. & Latency & PSNR  & SSIM   & FVD  \\
\hline
Wan 2.1     & 60M          & oom          & 32.16 & 0.8856 & 23.9 \\
Ours        & 14M         & 80ms           & 31.71 & 0.8788 & 27.1 \\
\bottomrule
\end{tabular}
\caption{Mobile efficient decoder for real-time decoding. }
\label{tab:efficientdecoder}
\end{table}

\subsection{Deployment Details and Speed Benchmark}

We provide a latency breakdown in \cref{tab:latencybreakdown}. 
In the case of mobile deployment, we solely use CLIP-ViT-L as the text encoder. 
We generate 3 latent frames for each chunk, and use a window size of 2 for KV cache. 
We deploy the model with Apple Coremltools \cite{coreml2021}, with the VAE decoder running on GPU and \methodAbbr $ $ running on neural engine to maximize resource utilization. 
We follow official practice \cite{coreml2021} to deploy \methodAbbr $ $ with 8-bit activation quantization and mixed-precision quantization for weights. 
Specifically, we keep sensitive layers (e.g., project in and out layers, text embedding layers) in 8-bit, but put most other layers in 4-bit. 
With deploy-time calibration, we observe minimal quality drop compared to the server BF16 model. 

\begin{table}[ht]
\small
\centering
\begin{tabular}{ccc}
\toprule
Model        & Latency (ms) & Steps \\
\hline
Text Encoder & 4            & 1     \\
\methodAbbr         & 260          & 4     \\
Decoder      & 80           & 1     \\
\hline
Total    &      1124        & FPS:10.7  \\
\bottomrule
\end{tabular}
\caption{Latency breakdown for generating a single chunk with 3 latent frames, which will be decoded to 12 pixel frames. }
\label{tab:latencybreakdown}
\end{table}

\begin{figure*}[ht]
    \centering
    \includegraphics[width=1.\linewidth]{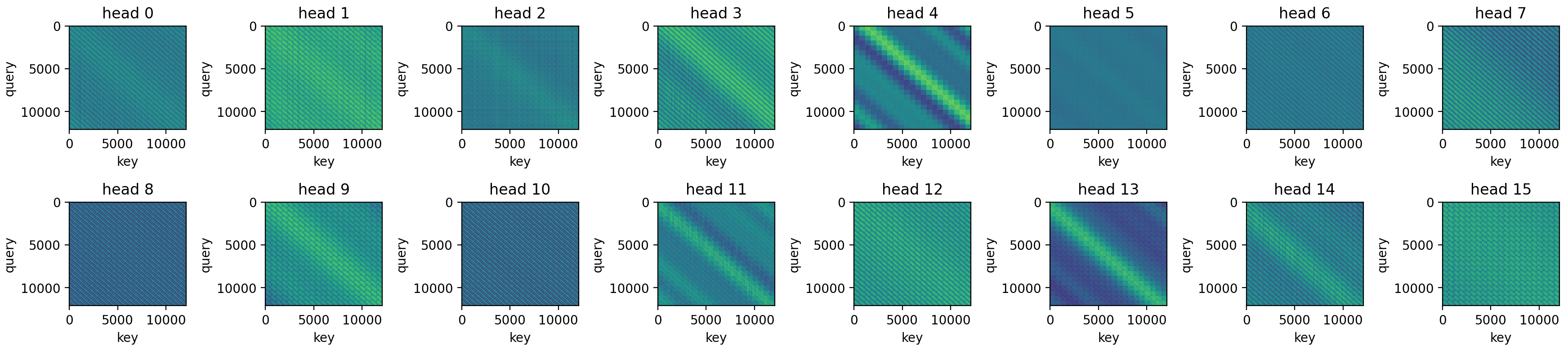}
    \caption{
    \textbf{Visualization of attention maps from our proposed linear attention.}
    }
    \label{fig:visual attn}
\end{figure*}

\section{User Study}

We perform a user study of text-to-video generation on 250 randomly sampled MovieGen VideoBench \cite{polyak2024movie} prompts. 
We compare the \methodAbbr-pretrained model, knowledge distilled model (KD), and autoregressive model (AR) with two baseline models, i.e., Wan2.1 1.3B \cite{wan2025wan} and LTX-0.9.5 \cite{LTX-video}. 
Human labelers are asked to pick the best from three anonymous and randomly shuffled videos. 
We instruct human labelers to focus on two metrics, 
(i) Text alignment, which evaluates whether the generated video follows the provided input prompt. 
(ii) Overall Quality, whether the generated video is visually pleasing, i.e., has complete generated object, meaningful and smooth motion, less flickering and artifacts, etc. 
Each metric is evaluated by at least 5 labelers and we take the average win rate. 
We find that with the detailed instruction, the variance of the picked results from different human labelers is generally low ($<3\%$ difference in win rate). 

As in \cref{tab:userstudy}, we show that our base model achieves on-par performance with the server-SOTA Wan2.1 1.3B model \cite{wan2025wan}, and outperforms server-efficient LTX \cite{LTX-video} by a large margin, demonstrating the superior performance of the proposed Sandwich Diffusion Transformer. 
With the subsequent 2-in-1 distillation, our KD full step model and streaming generation model achieves even higher win rate. 
We are the first to demonstrate high-quality streaming video generation but with mobile budget. 

\begin{table}[ht]
\small
\centering
\begin{tabular}{cccc}
\toprule
Model:          & \methodAbbr-Pre. & Wan2.1-1.3B & LTX-0.9.5 \\
\hline
Text Alignment  & 49.80\%   & 47.39\%     & 2.81\%    \\
Overall Quality & 46.99\%   & 43.37\%     & 9.64\%    \\
\hline
\hline
Model:          & \methodAbbr-KD   & Wan2.1-1.3B & LTX-0.9.5 \\
\hline
Text Alignment  & 61.45\%   & 36.66\%     & 1.89\%    \\
Overall Quality & 54.30\%   & 36.29\%     & 9.41\%    \\
\hline
\hline
Model:          & \methodAbbr-AR   & Wan2.1-1.3B & LTX-0.9.5 \\
\hline
Text Alignment  & 60.08\%   & 37.48\%     & 2.44\%    \\
Overall Quality & 57.39\%   & 35.62\%     & 6.99\%    \\
\bottomrule
\end{tabular}
\caption{User study on 250 Randomly Selected Prompts from MovieGen VideoBench \cite{polyak2024movie}. We show the win rate of each model. }
\label{tab:userstudy}
\end{table}

\section{Analysis of Linear Attention}

We visualize the attention maps produced by our linear-attention module. As shown in \cref{fig:visual attn}, the heads exhibit clear specialization: some emphasize temporal dynamics (e.g., heads 12 and 15), while others focus on spatial structure (e.g., heads 4 and 13), consistent with prior observations for self-attention \cite{xi2025sparse}. 
%

Besides, several heads capture global context (e.g., heads 2 and 5). In conjunction with the local path, this combination enables our model to learn global context and local detail simultaneously.
Consistent with this, our full model consistently outperforms the Local path-only variant across all evaluation metrics as illustrated in \cref{tab:efftattn_ablation_abs_delta}, confirming the importance of combining global and local.



\section{More Visualizations}
Beyond the horizontal results reported in the main paper, we include the comparisons of vertical videos that demonstrate the effectiveness of \methodAbbr.
As shown in \cref{fig: more visual1,fig: more visual2,fig: more visual3}, \methodAbbr~variants achieves stronger text–video alignment and richer visual details.
We include more videos and comparisons in the media file.


\begin{figure*}[ht]
    \centering
    \includegraphics[width=0.6\linewidth]{assets/Group09.pdf}
    \caption{
    \textbf{\textbf{Qualitative comparison on vertical videos.}}
    }
    \label{fig: more visual1}
\end{figure*}
\clearpage

\begin{figure*}[ht]
    \centering
    \includegraphics[width=0.6\linewidth]{assets/Group10.pdf}
    \caption{
    \textbf{\textbf{Qualitative comparison on vertical videos.}}
    }
    \label{fig: more visual2}
\end{figure*}
\clearpage

\begin{figure*}[ht]
    \centering
    \includegraphics[width=0.6\linewidth]{assets/Group11.pdf}
    \caption{
    \textbf{\textbf{Qualitative comparison on vertical videos.}}
    }
    \label{fig: more visual3}
\end{figure*}


\end{document}